\documentclass[sigconf]{acmart}
\usepackage{multirow}
\usepackage{float}
\usepackage{graphicx}
\usepackage{amsmath}
\usepackage{algorithmic}
\AtBeginDocument{%
  \providecommand\BibTeX{{%
    \normalfont B\kern-0.5em{\scshape i\kern-0.25em b}\kern-0.8em\TeX}}}

\setcopyright{acmcopyright}
\copyrightyear{2021}
\acmYear{2021}
\setcopyright{rightsretained}
\acmConference[MM '21]{Proceedings of the 29th ACM International Conference on Multimedia}{October 20--24, 2021}{Virtual Event, China}
\acmBooktitle{Proceedings of the 29th ACM International Conference on Multimedia (MM '21), October 20--24, 2021, Virtual Event, China}\acmDOI{10.1145/3474085.3478561}
\acmISBN{978-1-4503-8651-7/21/10}

\begin{document}
\fancyhead{}
\title{A Picture is Worth a Thousand Words: A Unified System for Diverse Captions and Rich Images Generation}

\author{Yupan Huang}
\authornote{Work done during an internship at Microsoft Research Asia.}
\affiliation{\institution{Sun Yat-sen University}}
\email{huangyp28@mail2.sysu.edu.cn}

\author{Bei Liu}
\affiliation{\institution{Microsoft Research Asia}}
\email{Bei.Liu@microsoft.com}

\author{Jianlong Fu}
\affiliation{\institution{Microsoft Research Asia}}
\email{jianf@microsoft.com}

\author{Yutong Lu}
\affiliation{\institution{Sun Yat-sen University}}
\email{luyutong@mail.sysu.edu.cn}

\renewcommand{\shortauthors}{Huang, et al.}

\begin{abstract}
A creative image-and-text generative AI system mimics humans' extraordinary abilities to provide users with diverse and comprehensive caption suggestions, as well as rich image creations.
In this work, we demonstrate such an AI creation system to produce both diverse captions and rich images.
When users imagine an image and associate it with multiple captions, our system paints a rich image to reflect all captions faithfully.
Likewise, when users upload an image, our system depicts it with multiple diverse captions.
We propose a unified multi-modal framework to achieve this goal.
Specifically, our framework jointly models image-and-text representations with a Transformer network, which supports rich image creation by accepting multiple captions as input.
We consider the relations among input captions to encourage diversity in training and adopt a non-autoregressive decoding strategy to enable real-time inference.
Based on these, our system supports both diverse captions and rich images generations.
Our code is available online\footnote{https://github.com/researchmm/generate-it}.

\end{abstract}




\maketitle

\begin{figure}[ht]
\begin{center}
   \includegraphics[width=1\linewidth]{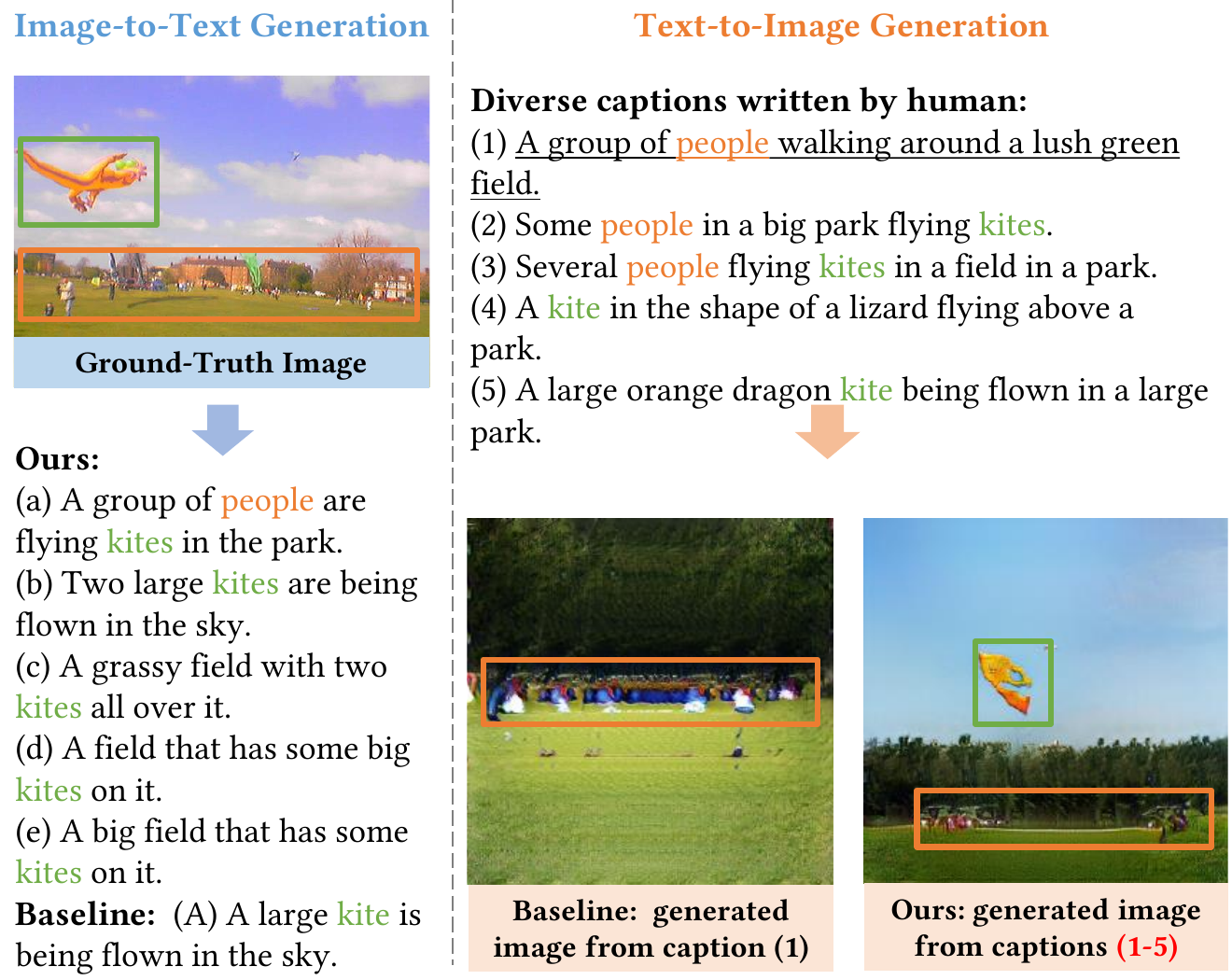}
\end{center}
   \caption{{
   Demonstration of our unified system to generate diverse captions from a rich image, and to generate a rich image from diverse captions.
   }}
\label{fig:intro}
\vspace{-4mm}
\end{figure}

\section{Introduction}

Image-and-text generative models mimic humans' extraordinary abilities to translate across real-scene images and natural language descriptions~\cite{li2020oscar,cho2020x}.
Image-to-text generation and text-to-image generation are bi-directional tasks.
Typical works formulate both tasks as a one-one mapping problem~\cite{anderson2018bottom,zhu2019dm}.
However, the restrictive bijective assumption has two issues.
First, it is ambiguous to explain an image with a single caption.
As the saying goes, ``A picture is worth a thousand words''.
An image is too rich and varied to be described by a caption with limited length.
Second, a single image-text pair cannot provide accurate and fine-grained alignment between objects in images and semantic units in captions.
This cross-modal alignment is important for image-and-text representation learning.

\begin{figure}[t]
\vspace{-3mm}
\begin{center}
   \includegraphics[width=1\linewidth]{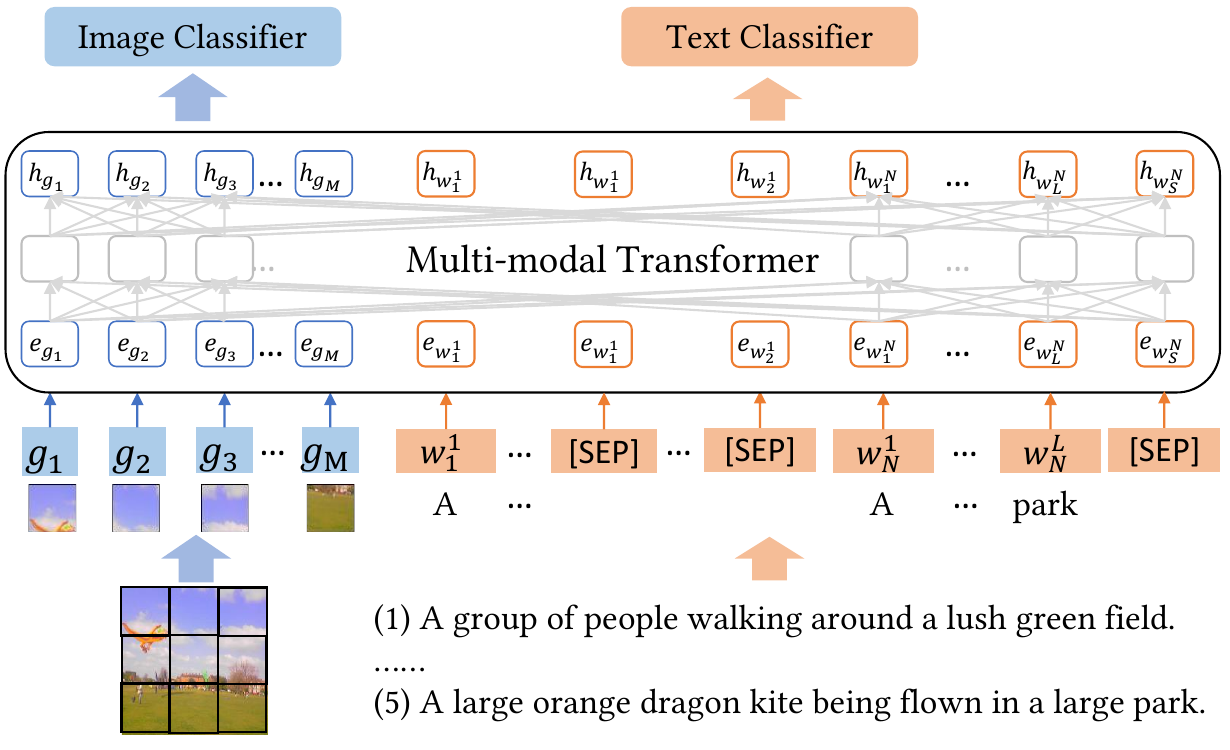}
\end{center}
   \caption{{Overview of our unified multi-modal Transformer.}}
\label{fig:transformer}
\vspace{-3mm}
\end{figure}

To tackle the above issues, we formulate bi-directional image-and-text generation tasks to align rich images and their corresponding multiple diverse captions.
Figure~\ref{fig:intro} demonstrates an example.
We get this inspiration from the diverse image captioning task, which aims to generate multiple captions to describe an image~\cite{wang2019describing}. However, we go one step further to formulate its bi-directional task of generating an image to reflect multiple captions.
Our tasks are different from image-to-paragraph generation~\cite{krause2017hierarchical} and vice versa, where our tasks focus on multiple independent captions instead of coherent captions.
We propose to generate both diverse captions and a rich image with a unified framework.
Specifically, our framework is based on a unified multi-modal Transformer to jointly model image-and-text representations.
To support diverse captions creation, we introduce an unlikelihood training objective to consider the relations among multiple input captions in training.
To support rich image creation, we construct a sequence of tokens from multiple captions as the input of the Transformer model.
Moreover, we adopt a non-autoregressive decoding strategy to enable real-time inference.
This has an application on auto-response tools to support diverse and inspiring suggestions from the AI system.
In summary, our contributions are three-fold:
\begin{itemize}
    \item Our work represents the first bi-directional image-and-text framework to generate both multiple diverse captions and rich images in the multimedia research community.
    \item We design a unified Transformer-based network and propose to consider relations among multiple input captions, which effectively improve the diversity of generated captions.
    \item We quantitatively and qualitatively demonstrate the effectiveness of our approach on the MSCOCO Captions dataset.
\end{itemize}

\section{Approach}
\subsection{Pipeline}
Our model mainly consists of a unified multi-modal Transformer as shown in Figure~\ref{fig:transformer}.
The Transformer~\cite{vaswani2017attention} is a decoder-only model in which image and text tokens are concatenated as inputs, and they can attend to each other in any one of the self-attention layers in the decoder.
The Transformer accepts a sequence of image and text representations as inputs, encodes them to contextualized vector representations, and outputs image and text tokens.
For text-to-image generation, we follow X-LXMERT \cite{cho2020x} to use an GAN-based image generator to convert image tokens to a real scene image.

\subsection{Image-and-Text Representations}
Given an image $X$, its caption set $Y^{1:N}=(Y^1,...,Y^N)$ is diverse, where $N=5$ for MSCOCO Captions dataset~\cite{lin2014microsoft}.
We denote the ground truth sequence of image tokens, text tokens of the $i$-th caption as $X=x_{1:M}$ and $Y^i=y^i_{1:L}$ respectively, where $M=8\times 8$ and $L=16$.
We shuffle $Y^{1:N}$ and concatenate them (separated by [SEP] tokens) as a sequence to serve as the inputs of our Transformer model.
We adopt the word embedding initialized with BERT for text representation following most Transformer-based models~\cite{devlin2019bert}.

We split an image into a sequence of uniform grid-level patches.
We extract the grid features with a Faster R-CNN~\cite{ren2015faster} object detector pre-trained on the Visual Genome dataset~\cite{krishna2017visual}. 
We use the original grid features as visual inputs for image-to-text generation tasks to reduce the loss of image information.
We adopt discrete clustering features of the original features to construct the ground-truth visual tokens as output prediction for text-to-image generation~\cite{cho2020x}.

\subsection{Training and Inference Strategy}
The general training objective for mutual image-and-text generation is to maximize the likelihood of the target ground-truth tokens given source context.
During training, we sample a masking ratio from a uniform prior distribution ([0,1]) and randomly mask the percentage of target tokens for prediction.
While in inference, we adopt a non-autoregressive sampling strategy (i.e., mask-predict-k strategy~\cite{ghazvininejad2019mask,deng2020length,cho2020x}).
So only a few sampling steps (e.g., 4) are needed to generate all target tokens, which enable real-time inference.

Existing diverse image captioning works assume $p(Y^{1:N}|X)=\prod_{i=1}^{N} p\left(Y^i |X \right)$ in the training phrase, which neglect the relations between caption set.
Thus the generated captions may be duplicated and lack diversity.
Instead, we propose to consider history generated captions when generating current caption $p(Y^{1:N}|X)=\prod_{i=1}^{N} p\left(Y^i |Y^{1:i-1}, X \right)$.
Specifically, in training phrase, we encourage diversity of the new generated caption from the history captions with a token-level unlikelihood objective~\cite{welleck2019neural}, and we propose to penalize repeating words in terms of word frequency.

\section{Evaluation}
We evaluate our method on the MSCOCO Captions dataset~\cite{lin2014microsoft} with popular Karpathy split~\cite{karpathy2015deep}.
Humans annotate the dataset with five captions for each image, which naturally suits our motivation.
We use Fréchet Inception Distance (FID) \cite{NIPS2017_8a1d6947} to measure the authenticity of the generated image.
The smaller the score value, the better the distribution of generated images matches with the distribution of real images.
For generated captions, we use {n-gram diversity (div-n)}~\cite{shetty2017speaking} to measure diversity and CIDEr-D~\cite{vedantam2015cider} to measure accuracy.

\begin{table}[]
\vspace{-3mm}
\centering
\caption{{Diverse image captioning results (\%) on MSCOCO Karpathy test split. The notation C denotes CIDEr-D score.}
}
\label{tab:ablation_diverse_image_caption}
\begin{tabular}{l|ll|ll}
\hline
\multicolumn{1}{c|}{\multirow{2}{*}{Training Method}} & \multicolumn{2}{c|}{Diversity Metrics} & \multicolumn{2}{c}{Accuracy Metrics} \\ \cline{2-5} 
\multicolumn{1}{c|}{} & \multicolumn{1}{c}{Div-1} & \multicolumn{1}{c|}{Div-2} & \multicolumn{1}{c}{Best C} & \multicolumn{1}{c}{Average C} \\ \hline
Baseline & 23.4 & 28.3 & \textbf{107.4} & \textbf{100.6} \\
Ours & \textbf{40.2} & \textbf{53.2} & 101.7 & 80.0 \\ \hline
\end{tabular}
\vspace{-3mm}
\end{table}

We demonstrate the effectiveness of our training strategy for diverse image captioning in Table~\ref{tab:ablation_diverse_image_caption}.
The diversity of the caption set generated by our approach significantly outperforms the baseline by a remarkable \textbf{16.8\%} and \textbf{24.9\%} absolute gains in Div-1 and Div-2 scores respectively.
Since our generated captions are diverse to describe an image from different aspects, they are not as accurate as the captions generated by the baseline in terms of CIDEr-D score.
However, our generated captions are fluent and meaningful as shown in Figure~\ref{fig:intro}.
For the text-to-image generation task, our design of generating images from diverse captions instead of from a single caption improves the FID score from \textbf{51.5} to \textbf{42.1} (lower is better).
The results quantitatively validate our system to enable effective diverse captions and rich images generation.

\bibliographystyle{ACM-Reference-Format}
\balance
\bibliography{base}

\end{document}